\def\year{2023}\relax
\documentclass[letterpaper]{article} 
\usepackage{aaai19}  
\usepackage{times}  
\usepackage{helvet}  
\usepackage{courier}  
\usepackage{url}  
\usepackage{graphicx}  
\usepackage{algorithm}
\usepackage{algorithmic}
\usepackage{amsmath}
\usepackage{amsfonts}
\usepackage{amssymb}

\def \GG {\mathbf{G}}

\DeclareMathOperator*{\argmin}{arg\,min}

\def \half {{\frac{1}{2}}}
\def \II {{\mathbf{I}}}

\def \MM {{\cal{M}}}

\def \EE {{\cal{E}}}
\def \CC {{\cal{C}}}
\def \HH {{\cal{H}}}
\def \PP {{\cal{P}}}

\def \EE {{\cal{E}}}
\def \HH {{\mathbf{H}}}
\def \AA {{\mathbf{A}}}
\def \XX {{\mathbf{X}}}
\def \BB {{\mathbf{B}}}
\def \ZZ {{\mathbf{Z}}}
\def \CC {{\mathbf{C}}}
\def \HHH {{\tilde {\mathbf{H}}}}

\def \MM {{\mathbf{M}}}
\def \MMM {{\tilde {\mathbf{M}}}}
\def \EE {{\mathbf{E}}}
\def \xx {\mathbf{x}}

\def \hh {\mathbf{h}}
\def \BB {\mathbf{B}}
\def \SS {\mathbf{S}}
\def \SSS {\hat{\SS}}
\def \DD {\mathbf{D}}
\def\Tr{{\mbox{Tr}}}

\newtheorem{lemma}{Lemma}

\begin{document}
%
\title{Robust Consensus Clustering and its Applications for Advertising Forecasting}

\author{Deguang Kong\thanks{This work was done during the authors were working at Yahoo Research.  The correspondence should be addressed to doogkong@gmail.com.  The views and conclusions contained in this document are those of the author(s) and should not be interpreted as representing the official policies, either expressed or implied, of any companies. }, 
Miao Lu, Konstantin Shmakov and Jian Yang \\
Yahoo Research,   San Jose, California, U.S.A, 94089 \\
doogkong@gmail.com, ml4ey@virginia.edu,  kshmakov@yahooinc.com,  jianyang@yahooinc.com \\
}
  
\maketitle
\begin{abstract}
Consensus clustering aggregates 
partitions in order to find a better fit by reconciling clustering results from different sources/executions. 
In practice, there exist noise and 
outliers in clustering task, which, however, may significantly degrade the performance. To address this issue, we propose a novel algorithm -- robust consensus clustering that
can find common ground truth among experts' opinions, which tends to be minimally affected by the bias caused by the outliers. In particular, we formalize the robust consensus clustering problem
as a constraint optimization problem,  and then derive an effective algorithm upon alternating direction method of multipliers (ADMM) with rigorous convergence guarantee. 
Our method outperforms the baselines on benchmarks. We apply the proposed method to the real-world advertising campaign segmentation and forecasting tasks using the proposed 
consensus clustering results based on the similarity computed via Kolmogorov-Smirnov Statistics. The accurate clustering result is helpful for building the advertiser profiles so as to perform the forecasting. 
\end{abstract}

\section{Introduction}

Consensus clustering reconciles clustering information from different sources (or different executions) to be better fit 
than the existing clustering results.  The consensus clustering results,
however can be biased due to different types of features, different
executions of algorithms, different definitions of distance
metrics and even different parameter settings, as is
sharply observed by existing studies~\cite{Jain:1999:DCR:331499.331504}.    All these factors may lead to disparity in clustering results. The major consensus clustering algorithms include Hypergraph
partitioning~\cite{Strehl:2003:CEK:944919.944935}, ~\cite{Fern:2004:SCE:1015330.1015414}, voting approach~\cite{btg038}, ~\cite{DBLP:journals/pami/RothLKB03}, mutual information ~\cite{DBLP:conf/icde/WangSC13}~\cite{DBLP:conf/sdm/TopchyJP04}, co-association approach~\cite{Xu:2005:SCA:2325810.2327433}, mixture model~\cite{DBLP:conf/sdm/TopchyJP04} ~\cite{Azimi:2009:ACE:1661445.1661603}, correlation consensus~\cite{Wang:2014:ECC:2647868.2654999}, ensemble clustering~\cite{Zhou:2015:LRC:2832747.2832822}, ~\cite{URN:NBN:fi:aalto-201610135020}, ~\cite{DBLP:conf/nips/LiuLY10}, ~\cite{7337436}, etc.

One key observation is that in consensus clustering ~\cite{Strehl:2003:CEK:944919.944935}, if there exist noise and outliers in one
source of features in any execution of algorithm, the clustering
result might be significantly affected due to the least square loss function used in most of the clustering methods (such
as k-means, Gaussian mixture model), because the errors are squared. Even worse, most of the time,
the users have little prior knowledge of noise, which makes the clustering result more
unstable and much harder to interpret with different initializations
and parameter settings. For example, if one information
source that we used for consensus clustering is not accurate,
when we align the consensus clustering results against
this ``inaccurate” source, we will suffer from these inaccurate
annotations. 
The
inaccurate common characteristics extracted from the samples due to the biased clustering results, are in fact, however,  less generalizable to those unseen
ones. 


To address these issues, this paper proposes a robust consensus
clustering schema that is minimally affected by the
outliers/noise. 
In particular, we combine multiple experts’ opinions on data clustering
using the robust $\ell_1$-loss function that aims to find the
maximum consistency on the experts' opinions with minimum
conflict. Our work can be viewed as an effective method
of data clustering from heterogeneous
information sources. The proposed method is practically
feasible because it is independent of parameter settings
of each clustering algorithm before aggregating the experts'
opinions. Driven by advertising applications, we apply the concensus clustering algorithm to cluster advertiser profiles~\cite{DBLP:conf/esa/MahdianW09} into different clusters so as to accurately perform performance (e.g.. Click, Conversions) forecasting. 

The main contribution of this paper is summarized
as follows.
\begin{itemize}
  \item To address the issue of consensus clustering performance degradation in existence of noise and outliers, we rigorously formulate the problem of robust consensus clustering 
  as an optimization problem using the robust loss function.   
  \item  To find the best solution for robust consensus clustering,  we develop an effective algorithm upon ADMM to derive the optimal solution, 
whose convergence can be rigorously proved. 

  \item  We experimentally
evaluate our proposed method on both benchmarks and real-world advertiser segmentation and forecasting tasks, and show that our
method is effective and efficient in producing the better clustering results. As an application, the proposed algorithm is applied for advertising forecasting tasks with more stable and trustful solutions.
\end{itemize}

\section{Robust consensus clustering Model}

{\bf Problem setting}
Assume we have $n$ data points,  each data point can generate features from different views (total view number=V).  For example, an image can generate features using different descriptors, such as SIFT\footnote{\texttt{https://en.wikipedia.org/wiki/\\Scale-invariant\_feature\_transform}}, HOG\footnote{\texttt{https://en.wikipedia.org/wiki/\\Histogram\_of\_oriented\_gradients}}, CNN features~\cite{DBLP:journals/corr/GirshickDDM13}.   More formally, 
let $\xx_i^v \in \Re^{p_v}$ be $v$-th ($1\leq v \leq V$) view/modal feature of a data point $i$, $p_v$ is the dimensionality of feature extracted from $v$-th view. 
 Consider all data points, $$\XX^v = [\xx_1^v, \xx^v_2, \cdots, \xx^v_n],$$
 where each data column vector is $\xx^v_i \in \Re^{p_v \times 1}$. Each data point has the ground-truth label  $y_i \in \Re^{K}$.  Simple concatenation of all data views gives
%
$$
\XX
= \left[ \begin{array}{c}
\XX^{1} \\
\XX^{2} \\
\vdots \\
\XX^{V} \\
\end{array} \right]
= [\xx_1, \xx_2, \cdots, \xx_n],
\quad
\xx_i
= \left( \begin{array}{c}
\xx^{1}_i \\
\xx^{2}_i \\
\vdots \\
\xx^{V}_i \\
\end{array} \right).
$$

Suppose we are given the clustering/partition results ($\PP$) for data points $ \{\XX^{1},  \XX^{2},  \cdots, \XX^{V} \}$ from $V$ different views, {\it i.e.},
$\PP = \{\GG^1, \GG^2, \cdots, \GG^V \}$, where $\GG^v \in \{0,1\}^{n\times K}$ (for K clusters) is the clustering result from view $v$.  The clusters/partitions assignment might be different for different views.  Let the connectivity matrix  (a.k.a co-association matrix) $\MM \in \{0,1\}^{n \times n}$,  where $M_{ij}(\GG^V)$ for view $v$ between data point $i,j$ is:
\begin{eqnarray}
\label{EQ:M}
M_{ij} (\GG^v ) = \left\{ {\begin{array}{*{20}c}
   {1; \;\;   
   \text{{\kern 1pt} {\kern 1pt} {\kern 1pt} (i,j){\kern 1pt} {\kern 1pt} belongs{\kern 1pt} {\kern 1pt} {\kern 1pt} to{\kern 1pt} {\kern 1pt} {\kern 1pt} the{\kern 1pt} {\kern 1pt} same{\kern 1pt} {\kern 1pt} {\kern 1pt} clustering}}  \\
   {0.   \;\;\;\;  \;\;\;\; \;\;\;\;  \;\;\;\;  \;\;\;\; \;\;\;\;  \;\;\;\; \;\;\;\; \;\;\;\; \;\;\;\; \;\;\;\;   
   \text{otherwise}
   }\\ 
\end{array}} \right.
\nonumber 
\end{eqnarray}

\subsubsection{Co-association consensus clustering} 

Standard consensus clustering  looks for a consensus clustering (based on majority agreement) $\GG^* \in \{0,1\}^{n \times K}$, such that $\GG^*$ is closest to the all the given partitions, {\it i.e.,}
\begin{eqnarray}
\label{EQ:J0}
\min_{\GG^*} J(\GG^*) =   \frac{1}{V} \sum_{v=1}^V \sum_{i=1}^n\sum_{j=1}^n D(M_{ij}(\GG^v) - M_{ij}(\GG^*)), 
\end{eqnarray}
where $D(.)$ is a distance function between the optimal solution and the solution from each view. Generally, least square loss is used for function $D(.)$: 
\begin{eqnarray}
\label{EQ:J01}
\min_{\GG^*} J(\GG^*) =   \frac{1}{V}  \sum_{v=1}^V \sum_{i=1}^n\sum_{j=1}^n  \Big(M_{ij}(\GG^v) - M_{ij}(\GG^*) \Big)^2, 
\end{eqnarray}

Notice that $M_{ij}(\GG^v) = {0,1}$, therefore, Eq.(\ref{EQ:J01}) can be equivalently written as: 
 \begin{eqnarray}
\label{EQ:J}
\min_{\GG^*} J_2(\GG^*) = \frac{1}{V}  \sum_{v=1}^V  \sum_{i=1}^n \sum_{j=1}^n |M_{ij}(\GG^v) - M_{ij}(\GG^*)| \nonumber \\
=   \frac{1}{V}  \sum_{v=1}^V \|M(\GG^v) - M(\GG^*)\|_{1},
\end{eqnarray}
with $\|X\|_{1} := \sum_{i,j} |X_{ij}|$. 

Let 
\begin{eqnarray}
\GG_{\ell_2} = \argmin_{\GG^*} J(\GG^*) \\
\GG_{\ell_1} = \argmin_{\GG^*} J_2(\GG^*). 
\end{eqnarray}
Clearly, 
\begin{eqnarray}
\GG_{\ell_2} = \GG_{\ell_1}. 
\end{eqnarray}

{\bf Analysis} 
Given little (or no) prior information of clustering result from each view, one natural way to compute the associate between data points $i$ and $j$ is 
to get the expected value of average association $\tilde{M} = [\tilde{M}_{ij}]$, {\it i.e.,}
\begin{eqnarray}
\tilde{M}_{ij} = \frac{1}{V} \sum_{v=1}^V M_{ij} (\GG^v).
\label{EQ:Mt}
\end{eqnarray}
Back to model of Eq.(\ref{EQ:J}),  the upper bound of $J_2$ is given by:
\begin{eqnarray}
J_2 \leq  \frac{1}{V}  \sum_{v=1}^V \| \MM(\GG^v) - \tilde{\MM} \|_1 + \| \tilde{\MM} - \MM(\GG^*)\|_1 \nonumber \\
= C +  \| \tilde{\MM} - \MM(\GG^*)\|_1 \nonumber \\
= C + J_3(\GG^*)
\label{EQ:J2}
\end{eqnarray}
where 
\begin{eqnarray}
J_3(\GG^*) =    \| \tilde{\MM} - \MM(\GG^*)\|_1  
\label{EQ:J3}
\end{eqnarray}
\begin{eqnarray}
C = \frac{1}{V}  \sum_{v=1}^V \| \MM(\GG^v) - \tilde{\MM} \|_1
\end{eqnarray}
Note the first term is a constant $C$ because it measures the average difference between the final clustering assignment and the average consensus association\footnote{The weighted average using attention can be adopted as well with learned weights.}  $\tilde{\MM}$. The smaller value of this term gives the closer distance between the final clustering assignment and current partition from a specific view.

On the other hand, 
\begin{eqnarray}
J_3 \leq \frac{1}{V}  \sum_{v=1}^V \Big(\| \MM(\GG^v) - \MM(\GG^*) \|_1 + \| \tilde{\MM} - \MM(\GG^v)\|_1 \Big) \nonumber \\
= J_2(\GG^*) + C 
\end{eqnarray}
This analysis indicates that the optimization of $J_2$ can be achieved by  $J_3$. 

{\bf Clustering indicator} 
One way to optimize $J_3$ of $\| \tilde{\MM} - \MM(\GG^*)\|_1$ of Eq.(\ref{EQ:J3}) is to assign the optimal clustering
indicator  $\GG \in \{0,1\}^{n \times K}$,  with the constraint that $ \sum_{k} G_{ik} = 1$, i.e.,  there is only one `1' in each row of $\GG$, and the rest are zeros. The connection between the averaged connectivity matrix $\MM(\GG^*)$ and $\GG$ is:
\begin{eqnarray}
\label{EQ:M=HH'}
\MM(\GG^*) = \GG^* \GG^{*{\intercal}}.
\end{eqnarray}
Therefore the objective of Eq.(\ref{EQ:J3}) becomes $J_4$, {\it i.e.,}
\begin{eqnarray}
& \min_{\GG} J_4(\GG) =    \| \tilde{\MM} - \GG \GG^{\intercal}   \|_1 \nonumber \\
& \;\; s.t., \;\;\; \GG \in \{0, 1\}^{n \times K}, \;\; \sum_k G_{ik} = 1
\label{EQ:J4}
\end{eqnarray}

\subsubsection{Relaxation  in continuous domain}

However, the major difficulty of solving Eq.(\ref{EQ:J4}) is that is involves the discrete clustering indicator $\GG$ and the algorithm is NP-complete~\cite{DBLP:journals/ijait/FilkovS04}. 
Also the objective involves non-smooth $\ell_1$ norm, which generally, is hard to handle.  
Thus, as most of the spectral clustering solvers, we do normalization on cluster indicators. In particular, we use $\HH \in \Re^{n \times K}$ to denote the new clustering indicator, {\it i.e.,}
$$\HH =  \GG(\GG^{\intercal} \GG)^{-\frac{1}{2}}.$$
Notice that 
\begin{eqnarray}
\label{EQ:H'H}
\GG^{\intercal} \GG = \left( {\begin{array}{*{20}c}
   {n_1 } & {} & {} & {} & {}  \\
   {} & {n_2 } & {} & {} & {}  \\
   {} & {} & {...} & {} & {}  \\
   {} & {} & {} & {...} & {}  \\
   {} & {} & {} & {} & {n_k }  \\
\end{array}} \right), 
\end{eqnarray}
where $n_k$ is the number of data points that falls in category $k$.  Let diagonal matrix $\DD \in \Re^{k \times k}$, {\it i.e.,}
$$\DD:= \GG^{\intercal} \GG.$$
then 
\begin{eqnarray}
\label{EQ:H-prop}
& \GG\GG^{\intercal} = \HH \DD \HH^{\intercal};   \\ \;\; 
& \HH^{\intercal} \HH =  \GG (\GG^ {\intercal} \GG)^{-1} = \II_k,
\end{eqnarray}
where $\II_k$ is an identity matrix with size kxk.
Therefore the objective function to be solved becomes:
\begin{eqnarray}
\label{EQ:M-HDH'}
&& \min_{\HH, \DD}  J_5 (\HH, \DD)= \|   \tilde{\MM} -   \HH \DD \HH^{\intercal} \|_1 \;\;   \nonumber \\
&&  s.t.,  \;\;\; \HH^{\intercal} \HH = \II_K,  \;\;  \DD \ge 0; \;\; \DD\;\; \text{is diagonal},
\end{eqnarray}
where $\GG = \HH \DD^{\half}$ can be obtained in continous domain. 

In practice, the solution obtained from $\hat{\MM}$  may not be accurate due to noise or outliers. 
For this reason, we term our method as ``robust'' because we use $\ell_1$ distance to measure the differences between the consensus optimal solution and the solution computed from each view. The errors are not squared.  Therefore, our model is capable of 
handling any noises/outliers in some view/execution in data clustering.

Further, we have:
\begin{lemma}
Set $\hat{\MM}$ to be the normalized pairwise similarity matrix $\hat{\SS} =\DD^{-\half} \SS \DD^{-\half}$ ($\SS \in \Re^{n \times n}$ and $S_{ij}$ is the similarity between data point $i$ and $j$),  using robust $\ell_1$-norm as the loss function, i.e., 
\begin{eqnarray}
&&  \min_{\HH} J_0(\HH) =    \| \hat{\SS}- \HH \HH^{\intercal}   \|_F^2 \nonumber \\
&&  \;\; s.t., \;\;\; \HH^{\intercal} \HH= \II_k. 
\label{EQ:J0}
\end{eqnarray}
This is identical to normalized cut spectral clustering~\cite{Shi:2000:NCI:351581.351611}.  
\end{lemma}

{\bf Proof } Recall that in standard normalized cut spectral clustering, the objective is: 
\begin{eqnarray}
&&  \min_{\HH}  \Tr (\HH^{\intercal}  (\II - \hat{\SS})\HH)  \nonumber \\
&&  \;\; s.t., \;\;\; \HH^{\intercal} \HH= \II_k,
\label{EQ:NC}
\end{eqnarray}
where $\hat{\SS} = \DD^{-\half} \SS \DD^{-\half}$. This is equivalent to optimizing:
\begin{eqnarray}
&&  \max_{\HH}  \Tr (\HH^{\intercal}  \hat{\SS} \HH) \nonumber \\
&&  \;\; s.t., \;\;\; \HH^{\intercal} \HH= \II_k.
\label{EQ:NC2}
\end{eqnarray}
Notice that 
\begin{eqnarray}
\nonumber 
 \| \hat{\SS}- \HH \HH^{\intercal}   \|_F^2 = \Tr(\SSS^{\intercal} \SSS+  \HH \HH^{\intercal} \HH^{\intercal} \HH - 2 \SSS^{\intercal} \HH \HH^{\intercal}),
\end{eqnarray}
 $\Tr(\HH \HH^{\intercal} \HH^{\intercal} \HH) = \text{const}$, and therefore 
$ \min_{\HH}\| \hat{\SS}- \HH \HH^{\intercal}   \|_F^2 $  is equivalent to  $\max_{\HH}  \Tr (\HH^{\intercal}  \hat{\SS} \HH)$.
This completes the proof. 

In this paper next, we discuss how to solve the  robust consensus clustering model of Eq.(\ref{EQ:M-HDH'}).

\section{Optimization Algorithm}

Eq.(\ref{EQ:M-HDH'}) seems difficult to solve since it involves non-smooth loss and high-order matrix optimization. We show how to apply alternating direction method of multipliers (ADMM) to solve it. ADMM method decomposes a large optimization problem by breaking them into smaller pieces, each of which are then easier to handle. ADMM 
combines the benefits of dual decomposition and augmented Lagrangian method for constrained optimization problem~\cite{constraint_optimization} and has been applied in many applications.  The problem solved by ADMM method usually has the following general forms\footnote{\small Here $\XX, \ZZ$ can be vectors or matrices.},  i.e.,
\begin{eqnarray}
\label{EQ:ADM}
&& \min_{\XX,\ZZ} f(\XX) + g(\ZZ), 
\;\;\;  s.t. \quad\quad \AA\XX + \BB\ZZ = \CC
\end{eqnarray}
After enforcing the Lagrangian multiplier while introducing more variables,  the problem can be solved  alternatively, i.e., 
\begin{eqnarray}
\ell(\XX,\ZZ,\mu) = f(\XX) + g(\ZZ) + \Omega^{\top} (\AA\XX + \BB\ZZ - \CC) 
\nonumber \\
+ \frac{\mu}{2}||\AA\XX + \BB\ZZ - \CC||_F^2,
\end{eqnarray}
\begin{eqnarray}
\label{EQ:update}
&& {\XX}^{t+1} := \argmin_{\XX} {\ell(\XX, \ZZ^t, \mu^t)}, \nonumber  \\
&& {\ZZ}^{t+1}:= \argmin_{\ZZ} {\ell(\XX^{t+1}, \ZZ, \mu^t)}, \nonumber \\
&& {\Omega}^{t+1}:= {\Omega}^t + \mu {(\AA \XX^{t+1} + \BB \XX^{t+1} -\CC)}. \nonumber \end{eqnarray}
and $\Omega$ is the augmented Lagrangian multiplier, and $\mu$ is the non-negative step size. 

\subsection{Optimization Algorithm to solve Eq.(\ref{EQ:M-HDH'})}

 According to ADMM algorithm, by imposing constraint variable $$\EE = \MMM-\HH \DD \HH^{^{\intercal}}, $$
  the problem of Eq.(\ref{EQ:M-HDH'}) is equivalent to solving,
\begin{eqnarray}
\label{EQ:E-H-D}
&& \min_{\EE, \; \HH, \; \DD}
\|\EE \|_1 ,  \nonumber \\
&&
s.t. \quad \quad \EE- (\MMM-\HH \DD \HH^{\intercal}) = 0; \nonumber  \\
&&
s.t., \HH^{^{\intercal}} \HHH= \II_K,  \;\;  \DD \ge 0; \;\; \DD\;\; \text{is diagonal}
\end{eqnarray}
In ADMM algorithm,
to solve $f(\EE)= \|\EE\|_1$, under the constraint 
$$h(\EE)=\EE- (\MMM-\HH \DD \HH^T),$$
 the ADMM function can be formulated as follows,
\begin{eqnarray}
\label{EQ:L(XYu)}
\ell(\EE, \HH, \DD,  \Omega,\mu) = f(\EE) + \Omega^{\top} h(\EE) + \frac{\mu}{2}||h(\EE)||_F^2. \nonumber
\end{eqnarray}
where Lagrange multiplier is $\Omega$ and $\mu$ is the penalty constant.
We solve a sequence of subproblems
%
\begin{eqnarray}
\label{EQ:L(YZEu)}
\min_{\EE, \HH, \DD}  \|\EE \|_1+  \Omega^{^\top} (\EE- (\MMM-\HH \DD \HH^ {^{\intercal}})) \nonumber \\
+ \frac{\mu}{2}   \|  \EE- (\MMM-\HH \DD \HH^{^{\intercal}}) \|_F^2.
\end{eqnarray}
%
with $\Omega$ and $\mu$ updated in a specified pattern: 
$$\small \Omega \leftarrow \Omega+ \mu(\EE- (\MMM-\HH \DD \HH^{^{\intercal}})),$$  
$$\small \mu \leftarrow \rho \mu.$$

To solve Eq.(\ref{EQ:L(YZEu)}), we search for optimal $\EE$, $\HH, \DD$ iteratively until the algorithm converges. 
Now we discuss how to solve $\EE$, $\HH$,  $\DD$ in each step. Alg.\ref{alg:robust-consensus} summarizes the complete algorithm.

\begin{algorithm}[t]
\small
 \caption{Solving robust consensus clustering model of  Eq.(\ref{EQ:M-HDH'})}
\label{alg:robust-consensus}
\hspace*{0.02in} {\bf Input:} clustering results from different views $\MMM$, parameter $\rho>1$. \\
\hspace*{0.02in} {\bf Output:} final clustering indicator $\HH$. \\
\hspace*{0.02in} {\bf Procedure:}
\begin{algorithmic}[1]
  \STATE Initialize $\EE^0$, $\HH^0, \DD^0$, $\Omega^0$, $\mu^0 > 0$, $t = 0$
  \WHILE{Not converge }
      \STATE Update $\EE$ via Eq.(\ref{EQ:Esol})
      \STATE Update $\HH$ via Eq.(\ref{EQ:Gsol})
      \STATE Update $\DD$ via Eq.(\ref{EQ:Dsol})
      \STATE $\mu^{t+1} := \rho \mu^{t}$
      \STATE  $\small \Omega:=  \Omega+ \mu(\EE- (\MMM-\HH \DD \HH^{^{\intercal}}))$
      \STATE $t := t+1$
   \ENDWHILE
\end{algorithmic}
\normalsize
\end{algorithm}

\subsubsection{Update $\EE$}

To update the error matrix $\EE$, we derive Eq.(\ref{EQ:Eupdate}) with
fixed $\HH, \DD$ and obtain the following form: 
%
\begin{eqnarray}
\label{EQ:Eupdate}
%
%
\min_{\EE} \ \frac{\mu}{2} ||\EE-\AA||_F^2 +  ||\EE||_1
\end{eqnarray}
where $$\AA = \HH\DD\HH^T - \MMM +\frac{\Omega}{\mu}, $$
It is well-known that the solution to the above
LASSO type problem~\cite{Wright:2009:RFR:1495801.1496037} is given by
\begin{eqnarray}
\label{EQ:Esol}
\EE_{ij} = sign(\AA_{ij}) \max(  |\AA_{ij}| - \frac{1}{\mu}, 0).
\end{eqnarray}

\subsubsection{Update $\HH, \DD$}

To update $\HH, \DD$ while  fixing $\EE$, we minimize the relevant part of Eq.(\ref{EQ:L(YZEu)})  
which is
%
\begin{eqnarray}
\label{EQ:Zupdate}
&& \min_{\HH, \DD}  \frac{\mu}{2} \|\BB - \HH\DD\HH^T\|_F^2, \nonumber \\ 
&& s.t., \HH^T \HH = \II_K,  \;\;  \DD \ge 0; \;\; \DD\;\; \text{is diagonal}
\end{eqnarray}
where
$$\BB = \MMM-\EE + \frac{\Omega}{\mu}.$$
%
%

It is easy to see the optimal solution $\HH$ is 
given by the $k$-largest eigenvectors of $\BB$,
i.e., $\HH =[\hh_1,\hh_2,\cdot \cdot \cdot, \hh_k]$,
\begin{equation}
\label{EQ:Gsol}
\BB  \hh_k = \lambda_k \hh_k ,
\end{equation}
where $\lambda_k$ is the associated eigen-value with respect to eigen-vector $\hh_k$, and
$\Lambda$ is a diagonal matrix, and $$\Lambda = diag([\lambda_1,\lambda_2, \cdot \cdot \cdot, \lambda_k]).$$
Then the optimal solution of $\DD$ is given by
\begin{eqnarray}
\DD = \Lambda.
\label{EQ:Dsol}
\end{eqnarray}
The completes the algorithm for solving Eq.(\ref{EQ:M-HDH'}). 

{\bf Time Complexity Analysis }
Since Eq.(\ref{EQ:M-HDH'}) is not a convex problem, in each iteration,
given $\mu$, $\Sigma$, $\rho$, the algorithm will find its local solution. The
convergence of ADMM algorithm has been proved and widely discussed in
~\cite{constraint_optimization}. 

The overall time cost for the algorithm depends on iterations and time cost for each variable updating. The computation of $\EE$ takes ${\cal O}(n k^2)$ time. The major burden is from the updating of $\DD$ and $\HH$, which requires computation of top $k$ eigen-vector of $\BB$,i.e., ${\cal O}(n^3)$ time. Overall, the cost of the algorithm 1 is ${\cal O}(T(n^3+n k^2))$, where $T$ is the iteration number before convergence.  To make the solution scalable for large-scale dataset, We can accelerate this via Hessenberg
reduction and QR iteration with multi-thread solver using GPU acceleration~\cite{Acceigen766} and also use divide-and-conquer~\cite{DBLP:conf/iccv/TalwalkarMMCJ13} to accelerate the execution.

\section{Experimental Results on Benchmarks}

\begin{table}
\centering
\small
\caption{\small dataset descriptions}
\label{tbl:dataset}
\begin{tabular}{c |c|c|c}
\hline \hline
datasets &  \#data points & \# feature & \#class  \\
\hline \hline 
CSTR  & 475 & 1000 & 4  \\
\hline
Glass & 214 & 9 & 7 \\
\hline
Ionosphere & 351 & 34 & 2 \\
\hline
Iris & 150 & 4 & 3 \\
\hline
Reuters & 2900 & 1000 & 10 \\
\hline
Soybean & 47 & 35 & 4 \\
\hline
Wine& 178 & 13 & 3 \\
\hline
Zoo & 101 & 18 & 7\\
\hline \hline
\end{tabular}
\end{table}
\begin{table*}
\centering
\small
\caption{\small Accuracy on 8 benchmark datasets}
\label{tbl:acc}
\begin{tabular}{c |c|c|c|c|c|c|c|c|c|c}
\hline \hline
datasets & k-means &  KC & CSPA & HPGA & NMFC & WC & L2CC & RCC (our method) & ES & CorC  \\
\hline \hline 
CSTR  &   0.45 &  0.38 & 0.50 & 0.62 & 0.56 & 0.64 & 0.61 & {\bf 0.65} & 0.64 & 0.61\\
\hline
Glass & 0.38 & 0.45 & 0.43 & 0.40 & 0.49 & 0.49 & 0.49 & {\bf 0.52} & 0.52 & 0.50 \\
\hline
Ionosphere & 0.70 & 0.71 & 0.68 & 0.52 & 0.71 & 0.71 & 0.71 & {\bf 0.72} & 0.71 & 0.70 \\
\hline
Iris & 0.83 & 0.72 & 0.86 & 0.69& {\bf 0.89} & {\bf 0.89} &  0.86 & {\bf 0.89} & 0.88 & 0.86\\
\hline
Reuters  & {\bf 0.45} & 0.44 & 0.43 & 0.44& 0.43 & 0.44& 0.43 & {\bf 0.45} & 0.44 & 0.43\\
\hline
Soybean & 0.72 & 0.82 & 0.70 & 0.81 & 0.89 & 0.91 &  0.76 & {\bf 1.00} & 0.98 & 0.95 \\
\hline
Wine& 0.68 & 0.68 & 0.69 & 0.52 & 0.70 & 0.72 & 0.65 & {\bf 0.96} & 0.84 & 0.89 \\
\hline
Zoo &  0.61 & 0.59 & 0.56 & 0.58 & 0.62 & 0.70& 0.80 & {\bf 0.84} & 0.79 & 0.82\\
\hline \hline
\end{tabular}
\end{table*}
\begin{table*}
\centering
\small
\caption{\small Clustering accuracy on three multi-view datasets}
\label{tbl:mv-acc}
\begin{tabular}{c |c|c|c|c|c|c|c|c|c}
\hline \hline
datasets & LBP & HOG  & GIST & Classemes & FC & MVKmean & AP & LCC & RCC (Our method)  \\
\hline \hline 
MSRC-v1 & 0.4731 & 0.6367 & 0.6283 & 0.5431 & 0.7423 & 0.7871 & 0.5369 & 0.7542 &  {\bf 0.8017}
\\
\hline
Caltech-7  & 0.5236 &  0.5561 & 0.5473 & 0.4983 & 0.6123 & 0.6640 & 0.5359 & 0.6643 & {\bf 0.6819} \\
\hline
Caltech-20 & 0.3378 & 0.3679 & 0.3925 & 0.3660 & 0.5489 & 0.5619 & 0.3421 & 0.6287 & {\bf 0.6374} \\
\hline \hline
\end{tabular}
\end{table*}

In order to validate the effectiveness of our method, we perform experiments on benchmark datasets. In particular, we use six datasets downloaded from UCI machine learning repository\footnote{\texttt{https://archive.ics.uci.edu/ml/datasets.html}},
including Glass, Ionosphere, Iris, Soybean, Wine, zoo. We also adopted two widely used text datasets for document clustering, including CSTR\footnote{\texttt{https://github.com/franrole/cclust\_package/\\tree/master/datasets}} and Reuters. The features we adopt are represented using vector space model after removing the stop words and unnecessary tags and headers.  In Reuter dataset, we use the ten most frequent categories. 

In our experiment, the consensus co-association matrix is obtained by running k-means algorithm for 40 times and make an average of results using different initializations. All datasets have the ground-truth of clustering labels. We also compare against the following baseline methods:
\begin{itemize}
\item k-means on original feature space
\item KC:  k-means on the consensus similarity matrix;
\item NMFCC~\cite{Li:2007:SCS:1441428.1442121}: NMF-based consensus clustering; 
\item CSPA~\cite{Domeniconi:2009:WCE:1460797.1460800}: cluster based similarity partition algorithm 
\item HPGA~\cite{Strehl:2003:CEK:944919.944935}:  Hypergraph partitioning algorithm 
\item WCC~\cite{DBLP:conf/sdm/LiD08}: weighted consensus clustering 
\item L2CC: using standard $\ell_2$ loss for solving the consensus clustering on the consensus matrix; 
\item RCC: proposed robust consensus clustering algorithm
\item Correlation Consensus (CorC)~\cite{Wang:2014:ECC:2647868.2654999}: Correlation
based consensus clustering by exploiting the
ranking of different views;
\item Ensemble Consensus(ES)~\cite{Zhou:2015:LRC:2832747.2832822}: ensemble consensus via exploiting Kullback-Leibler divergence of different aspects.
\end{itemize}

{\bf Evaluation metrics} As the other clustering tasks, we use the  \emph{clustering accuracy} as the metric to evaluate the performance of our algorithms. 
\textbf{Clustering accuracy(ACC) }is defined as,
\begin{equation}
ACC = \frac{{\sum\nolimits_{i = 1}^n {\delta ({l_i},map({c_i}))} }}
{n},
\end{equation}
where $l_i$ is the true class label and $c_i$ is the obtained
cluster label of $x_i$, map(.) is the best mapping function, $\delta
(x,y)$ is the delta function where $\delta (x,y)=1$ if $x = y $, and
$\delta (x,y)=0 $ otherwise. The mapping function map(.) matches the
true class label and the obtained clustering label, where the best
mapping is solved by Hungarian algorithm. The larger, the better. 

{\bf Experiment result analysis} 
We report clustering results  in Table.~\ref{tbl:acc}. Clearly, our method is very robust and outperforms the other methods. 
The accuracy gain is very significant especially on dataset \texttt{Soybean, Wine} and \texttt{zoo}. In some cases,
the clustering result from one execution might be severely biased. 
This can be addressed by running clustering multiple times
to reduce the variance. The Correlation Consensus algorithm~\cite{Wang:2014:ECC:2647868.2654999} is designed for multi-modal data, which did not
show very promising performance for the multiple execution
of datasets with different initializations.


\subsection{Experiment results on multi-view datasets}

{\bf Dataset} Our algorithm can be easily extended for processing multi-view data when $ \tilde{\MM}$ 
of Eq.(\ref{EQ:M-HDH'}) is computed using multi-view features. We evaluate this over several 
public datasets: Caltech101~\cite{Fei-Fei:2007:LGV:1235884.1235969} and Microsoft
Research Cambridge Volume 1 (MSRCV1)~\cite{1541329}. 
MSRCV1 has images from 7 classes, i.e., airplane, bicycle,
building, car, cow, face, and tree. Each class has 30 images.
Caltech-101 is an image dataset of 8677 images and 101 object
classes. We follow the same experimental procedure as~\cite{4408853} and extract 7 and 20 classes for each experimental setup.  We extract 1984-dimensional LBP~\cite{OJALA199651} features, 4000-dimensional GIST~\cite{Oliva:2001:MSS:598425.598462} features, 768-dimensional HOG~\cite{1467360} features, 2659-dimensional
classemes~\cite{Torresani:2010:EOC:1886063.1886122} features from each image.

{\bf Comparison results} We compare our methods against clustering using each view of features, i.e., LBP, HOG, GIST and Classemes,  respectively.
We also compare several multi-view clustering methods: Spectral clustering using simple multi-view feature concatenation (denoted as FC), multi-view k-means (denoted as MVKmean)~\cite{Cai:2013:MKC:2540128.2540503}, affinity propagation using multi-view features ( denoted as AP)~\cite{4408853}, low-rank consensus clustering (denoted as LCC)~\cite{ijcai2017-396}.  Table.~\ref{tbl:mv-acc} presents the clustering accuracy results, which demonstrates the superiority of using our method for solving multi-view clusering problems. Moreover, our method is flexible for incorporating any view of features after properly feature extraction. Essentially, our method learns the low-rank subspace using eigen-decomposition using co-association matrix from multi-view data, which plays the similar role as those in~\cite{ijcai2017-396}, ~\cite{Cai:2013:MKC:2540128.2540503}.  

\section{Applications in Advertising Segmentation}


Computational Advertising refers to serving the most relevant advertisement (ads for short) matching to the particular context in internet webs.  
One import problem is to  \emph{segment advertisers into different segmentations and provide personalized service for targeting selection (i.e., provide the best target audiences) and forecasting (i.e., predict clicks and conversions for advertisers in winning auctions). }


\begin{figure}[t]
	\centering
	\includegraphics[height=1.5in,width=0.5\textwidth]{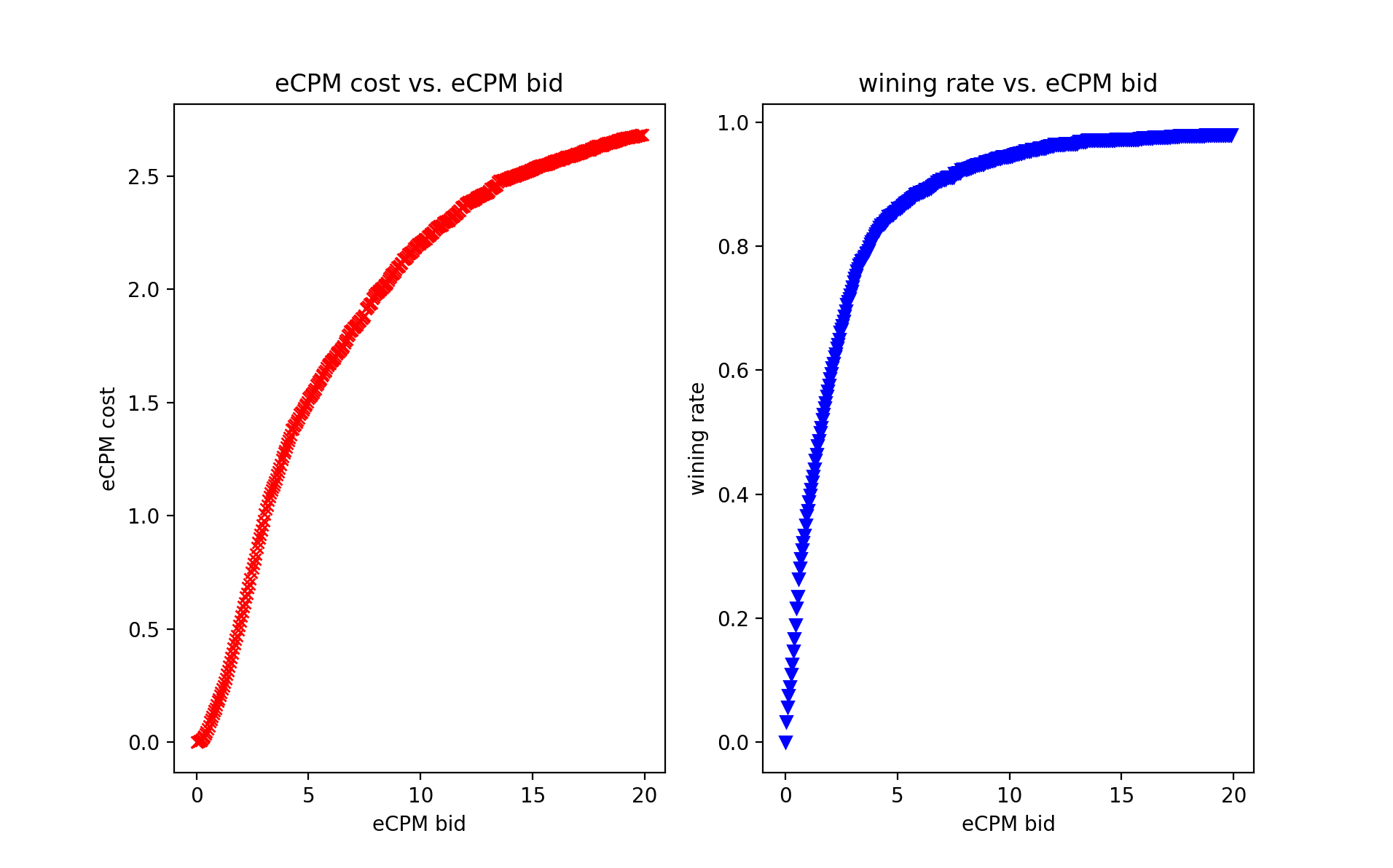}
	\caption{\small (left) eCPM cost \texttt{c.d.f} vs. eCPM bid; 
(right) win rate \texttt{c.d.f} vs. eCPM bid.\\ \\}
	\label{fig:BL}
\end{figure}


To cluster advertisers into different segmentations, we have three views of information: (i) advertisers' textual descriptions; (ii) advertisers' win-rate cumulative distribution function (i.e., \texttt{c.d.f} for short); (iii) advertisers' eCPM cost \texttt{c.d.f}. In advertising, win rate is a percentage metric in programmatic media marketing that measures the number of impressions won over the number of impressions bid, and win rate cumulative distribution function (\texttt{c.d.f}) measures the percentages of impressions won blow certain bids.  eCPM cost is the effective cost per thousand impressions given the bid, and eCPM cost cumulative distribution function (\texttt{c.d.f}) measures the cost of impressions below certain bids. Fig.\ref{fig:BL} give an example of advertisers' win-rate \texttt{c.d.f} and eCPM cost \texttt{c.d.f} (a.k.a  bid landscape model that allows one to estimate performance  of advertising campaigns under different bidding scenarios).  We extract advertisers' textual description for clustering purpose.

{\bf Connection to the proposed robust clustering method} Our method consists of two key steps: (i) generating clustering results using each view information and obtain the co-association matrix of $\tilde{M}$ of Eq.(\ref{EQ:Mt}) 
(ii) using the robust consensus clustering model of Eq.(\ref{EQ:M-HDH'}) to generate the consensus clustering results.  

{\bf Similarity Computation using KS Statistics }
Given the pairwise similarity in each view, a natural way is to segment different them into different disjoint $K$ subsets where each subset denotes a clustering using graph cut type algorithm (e.g., Normalized Cut).  In order to generate the clustering results using win-rate \texttt{c.d.f}  and eCPM cost  \texttt{c.d.f}, we refer to Kolmogorov- Smirnov (KS for short) statistics~\cite{kolmogorov1933sulla},~\cite{smirnov1948},  which offers a \emph{nonparametric} way to quantify the distances between the empirical distribution functions. 
The traditional way is to compare the mean or median of two distributions with parametric tests (using Z-test or t-test), or non-parametric tests (Mann-Whitney or Kruskal-Wallis test). However, these methods only consider the domain differences of two distributions, not the scale. Moreover, some parametric methods impose strong normal distribution assumption, which is not realistic for ad campaigns. 

Suppose we have independent identical distributed (\texttt{i.i.d}) samples $X_1, \cdots, X_m$ of size $m$ with cumulative distribution function (\texttt{c.d.f}) $F(x)$ and the second \texttt{i.i.d} sample $Y_1,...,Y_n$ of size $n$ with \texttt{c.d.f} $G(x)$. One wants to test the null hypothesis ($H_0: F=G$) vs. the alternative ($H_1: F \neq G$). $F_m(x)$ and $G_n(x)$ are the corresponding \texttt{c.d.fs}, where
\begin{eqnarray}
F_m(x)=\frac{1}{m}\sum_{i=1}^mI(X_i\leq x), \;\;\;
G_n(x)=\frac{1}{n}\sum_{i=1}^nI(Y_i\leq x).
\nonumber
\label{eq:ecdf}
\end{eqnarray}
The KS test statistic is defined as the maximum value of the absolute difference between two \texttt{c.d.f.s}, {\it i.e.,}
\begin{equation}
D_{mn}=\sqrt{\frac{mn}{m+n}}\sup_x |F_m(x)-G_n(x)|
\label{eq:ks}
\end{equation}
The distribution of KS test statistic doesn't depend on the known distribution. In addition, it converge to KS distribution, {\it i.e.}
\begin{equation}
\lim_{m,n\rightarrow \infty}P(D_{mn}\leq z)=1-2\sum_{i=1}^\infty(-1)^{i-1}exp(-2i^2z^2)
\label{eq:ks dist}
\end{equation}
This nice property enables the wide application of KS-statistics for solving real-world problems.  In the context of advertiser clustering, 
let $f_{m}(i)$ be the winning rate function \texttt{c.d.f} for advertiser $i$ generated from $m$ observations, and $f_{n}(j)$ be the win rate function \texttt{c.d.f} for advertiser $j$ generated from $n$ observations,  then distance $dist(i,j)$ between $i$ and $j$ is defined as: 
\begin{equation}
dist(i,j|m,n)=\sqrt{\frac{mn}{m+n}}\sup_x |f_m(BF_i)-f_n(BF_j)|.
\label{eq:ks2}
\end{equation}
Therefore the similarity (denoted as $S^{\text{win}}_{ij}$) of pairwise advertisers $i, j$ is computed based on 
\begin{eqnarray}
\label{EQ:sim}
S^{\text{win}}_{ij}=\left\{\begin{matrix}
1; \;\;\;\text{if} \;\;\; dist(i,j|m,n) \leq C_{\alpha}  \\ 
0; \;\;\;\; \;\;\;\;\;\;\;\;\;\;\;\;\; \text{otherwise},
\end{matrix}\right.
\end{eqnarray}
where $C_{\alpha}$ is the confidence value set corresponding to $p$-value. In particular, if one chooses a critical value $C_{\alpha}$ such that $$\Pr(dist(i,j|m,n) > C_{\alpha}) = \alpha,$$ then a band of width $C_{\alpha}$ around the distribution of $dist(i,j|m,n)$  will entirely contain
the estimated value with probability $(1-\alpha)$. eCPM cost similarity is computed using the same way. 

\vspace{-2mm}
\subsection{Experimental Results}

The ad campaign data are collected from a major web portal. In this study, we consider 1,268 advertisers over a week.  As is introduced 
in Kolmogorov-Smirnov statistics, we set $p$-value = $\{0.01, 0.05, 0.1\}$ ($p=0.01$ lower bound and $p=0.1$ upper bound). 
Table~\ref{tbl:sim} shows the clustering number using the win rate \texttt{c.d.f} and eCPM cost  \texttt{c.d.f} features respectively. 
Clearly, when $p$-value increases, the critical $C_{\alpha}$-value for determination of similar campaigns decreases because there are 
smaller chances for advertisers falling in the range of $P(dist(i,j|m,n))>C_{\alpha}$ using Eq.(\ref{EQ:sim}).   

{\small
\vspace{+1mm}
\begin{table}[ht]{}
\centering
\caption{ \# of clusterings at different $p$}
\label{tbl:sim}
\begin{tabular}{c |r| r|r}
\hline  \hline
Category &  $p=0.01$ &  $p = 0.05$    &  $p=0.1$  \\
\hline \hline
win rate  	& 65 & 47  & 32 \\
\hline
eCPM&  73 & 51 & 40 \\
\hline \hline
\end{tabular}
\end{table}
}

{\bf Consistency of clustering } 
 For any pairwise advertisers, if both win rate \texttt{c.d.f} and eCPM cost \texttt{c.d.f} cluster them to the same group (or the different group), then they are viewed as true consistency, 
otherwise, if one method clusters them in the same group while the other clusters them in different groups, then viewed as inconsistency (FPN).  %
More mathematically,  let $C_i, C_j$ be the cluster label obtained using eCPM cost for $i, j$ and let $\ell_i, \ell_j$ be the cluster using win rate for $i$ and $j$ , then 
{\small
\begin{eqnarray}
\label{EQ:tp_ecpm}
 \small{TPN} =  \sum_{i < j} \Big[  \Big( I_{C_i = C_j} \Big) \bigcap  \Big(I_{\ell_i = \ell_j} \Big) \Big]   
  \bigcup    \Big[ \Big(I_{C_i \neq C_j} \Big) \bigcap  \Big(I_{\ell_i \neq \ell_j} \Big) \Big] \nonumber 
\end{eqnarray}
\begin{eqnarray}
\label{EQ:fn_ecpm}
 \small{FPN} =  \sum_{i < j} \Big[ \Big(I_{C_i = C_j} \Big) \bigcap  \Big(I_{\ell_i \neq \ell_j} \Big)  \Big]  
 \bigcup \Big[  \Big(I_{C_i \neq C_j} \Big)  \bigcap \Big(I_{\ell_i = \ell_j} \Big) \Big]  \nonumber 
\end{eqnarray}
}
{\small
\begin{table}[ht]
\centering
\caption{Consistency of clustering using win rate \texttt{c.d.f} and eCPM cost \texttt{c.d.f} }
\label{tbl:cluster}
\begin{tabular}{r|r}
\hline \hline
Category &   Percentage \%  \\
\hline \hline
TPN & 74.21\% \\
FPN & 25.79\%  \\
\hline\hline
\end{tabular}
\end{table}
}

Table~\ref{tbl:cluster} shows the consensus clustering results using these two views of features. Clearly, around three fourth advertisers share very similar results even using different features. 
{\small
\vspace{+2mm}
\begin{table}[th]
\centering
\small
\caption{\small Forecast impression error (the smaller, the better)}
\label{tbl:imp_fore}
\begin{tabular}{c |c}
\hline \hline
Feature type & \# forecasting error  \\
\hline \hline 
win c.d.f &  18.10\%\\
eCPM c.d.f & 20.31\%\\
textual &  30.98\%\\
consensus clustering & 17.87\% \\
propose method & {\bf 15.96\%} \\
\hline\hline
\end{tabular}
\end{table}
}

{\small
\begin{table}[th]
\centering
\small
\caption{\small Forecast spend error (the smaller, the better)}
\label{tbl:spend_fore}
\begin{tabular}{c |c}
\hline \hline
Feature type & \# forecasting error  \\
\hline \hline 
win c.d.f &  12.34\%\\
eCPM c.d.f & 13.56\%\\
textual &  25.78\%\\
consensus clustering & 12.78\% \\
propose method & {\bf 9.64\%}\\
\hline \hline
\end{tabular}
\end{table}
}

{\bf Clustering performance } Given the fact that we do not have ground-truth for advertiser clustering results, we cannot directly evaluate the performance of our method. Instead, we view the robust consensus clustering result as 
the advertiser segmentation result, and use it to re-generate the eCPM cost and win-rate \texttt{c.d.ds} using all the information from the same advertiser clusters. With the updated eCPM cost~\cite{Cui:2011:BLF:2020408.2020454} and win-rate \texttt{c.d.f.s}, 
we forecast the clicks and spend, and compute the relative percentage errors to validate the performance of updated eCPM cost and win-rate distributions.  In particular, if the clustering result is better, then 
the re-generated eCPM and win-rate distributions are more accurate.
\begin{eqnarray}
\texttt{ \text{\# Impression} = \text{Total supply}} \times  \texttt{\text{win-rate} } \nonumber \\
\texttt{\text{\# Spend } = \text{\# Impression}}  \times \texttt{ \text{ eCPM cost}} \nonumber 
\end{eqnarray}
which is to say, the forecasted impression (\text{\# Impression}) is equal to the total number of supplies by applying the win-rate, and the forecasted spend is equal to the forecasted impressions multiplied by eCPM cost 
for per impression. The accurate estimation of \text{win-rate} and \text{eCPM} will make the forecasted impressions and spend more accurate. 

Therefore, we re-generate the  eCPM and win-rate distributions using the following several consensus clustering results:
(i) only ads textual;  (ii) only win-rate; (iii) only eCPM cost;  (iv) proposed robust consensus clustering;  (v) consensus clustering using $\ell_2$ distance; and compare their performances.  The mean of absolute percentage error is defined as:
$
MAPE = \frac{1}{n}\sum_{i=1}^n  \frac{|y_i - \hat{ y}_i |} {y_i},
$
where $y_i$ is the ground-truth for advertiser $i$ and $\hat{y}_i$ is the forecasted impression (or spend) for advertiser $i$. The smaller of these values, the better performance of clustering. 
Tables~\ref{tbl:imp_fore}, ~\ref{tbl:spend_fore} show the forecasted impression and spend errors, respectively.  Thanks to the proposed robust consensus  clustering algorithm, 
the  forecasting error has been dropped significantly for  both forecasted clicks and spend.  

\section{Conclusion}

This paper presents a novel approach for consensus clustering. The new clustering objective is more robust to noise and outliers. 
We formulate the problem rigorously
and show that the optimal solution can be derived using ADMM algorithm.  We apply our method 
to solve real-world advertiser segmentation problem, where the consensus clustering produces better forecasting 
results.


\bibliographystyle{aaai}
\small
\bibliography{rc}

\end{document}